\def\year{2021}\relax
\documentclass[letterpaper]{article} % DO NOT CHANGE THIS
\usepackage{aaai21}  % DO NOT CHANGE THIS
\usepackage{times}  % DO NOT CHANGE THIS
\usepackage{helvet} % DO NOT CHANGE THIS
\usepackage{courier}  % DO NOT CHANGE THIS
\usepackage[hyphens]{url}  % DO NOT CHANGE THIS
\usepackage{graphicx} % DO NOT CHANGE THIS
\usepackage{natbib}  % DO NOT CHANGE THIS AND DO NOT ADD ANY OPTIONS TO IT
\usepackage{caption} % DO NOT CHANGE THIS AND DO NOT ADD ANY OPTIONS TO IT
\usepackage{subfigure}
\usepackage[switch]{lineno}
\title{DeepWriteSYN: On-Line Handwriting Synthesis \\ via Deep Short-Term Representations}
\author{Ruben Tolosana, \textsuperscript{\rm 1}
    Paula Delgado-Santos, \textsuperscript{\rm 1}
    Andres Perez-Uribe,\textsuperscript{\rm 2}\\
    Ruben Vera-Rodriguez,\textsuperscript{\rm 1}
    Julian Fierrez,\textsuperscript{\rm 1}
    Aythami Morales \textsuperscript{\rm 1}
	\\
}
\affiliations{
    \textsuperscript{\rm 1}Biometrics and Data Pattern Analytics Lab, Universidad Autonoma de Madrid \\ Email: (ruben.tolosana, ruben.vera, julian.fierrez, aythami.morales)@uam.es, paula.delgadod@estudiante.uam.es
    
        \textsuperscript{\rm 2}University of Applied Sciences Western Switzerland (HEIG-VD) \\ Email: Andres.Perez-uribe@heig-vd.ch
}
\begin{document}
%\linenumbers  %
\maketitle

\begin{abstract}
This study proposes DeepWriteSYN, a novel on-line handwriting synthesis approach via deep short-term representations. It comprises two modules: \textit{i)} an optional and interchangeable temporal segmentation, which divides the handwriting into short-time segments consisting of individual or multiple concatenated strokes; and \textit{ii)} the on-line synthesis of those short-time handwriting segments, which is based on a sequence-to-sequence Variational Autoencoder (VAE). The main advantages of the proposed approach are that the synthesis is carried out in short-time segments (that can run from a character fraction to full characters) and that the VAE can be trained on a configurable handwriting dataset. These two properties give a lot of flexibility to our synthesiser, e.g., as shown in our experiments, DeepWriteSYN can generate realistic handwriting variations of a given handwritten structure corresponding to the natural variation within a given population or a given subject. These two cases are developed experimentally for individual digits and handwriting signatures, respectively, achieving in both cases remarkable results.

Also, we provide experimental results for the task of on-line signature verification showing the high potential of DeepWriteSYN to improve significantly one-shot learning scenarios. To the best of our knowledge, this is the first synthesis approach capable of generating realistic on-line handwriting in the short term (including handwritten signatures) via deep learning. This can be very useful as a module toward long-term realistic handwriting generation either completely synthetic or as natural variation of given handwriting samples.

%To the best of our knowledge, this is the first synthesis approach proposed for on-line handwritten signature via deep learning. 
\end{abstract}

\noindent The availability of large-scale public databases, together with the remarkable progress of deep learning, have led to the achievement of very impressive results in many different fields. However, there are still applications and scenarios in which data are scarce, achieving in general poor results. These scenarios are usually known in the literature as zero-, one-, and few-shot learning~\cite{wang2019survey,10.1145/3386252}. 

Different approaches have been proposed in response to the lack of data, being the synthesis of artificial data one of the main research lines followed nowadays. This is of special relevance in image-based scenarios due to the popular Generative Adversarial Networks (GAN) and Transfer Learning techniques~\cite{goodfellow2014generative,tan2018survey}. However, when the scenarios considered are based on time sequences, and not images, the synthesis of new data seems to be a more challenging task (or at least less developed). In this study we focus on the synthesis of time-sequences data for on-line handwriting applications such as signature and digit recognition.

\begin{figure*}[t]
\begin{center}
   \includegraphics[width=0.90\linewidth]{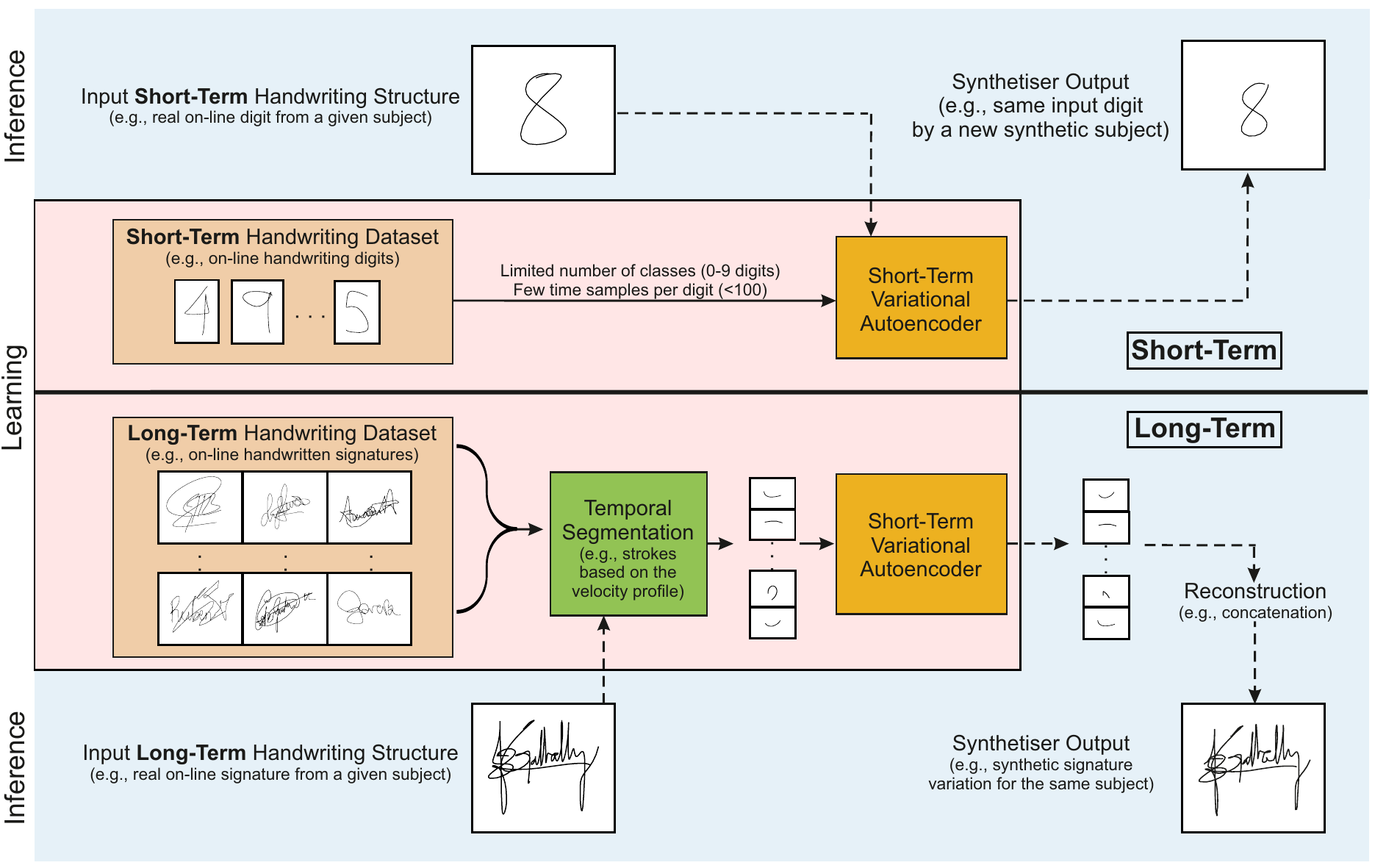}
\end{center}
   \caption{Diagram of DeepWriteSYN, the proposed on-line handwriting synthesis approach via deep short-term representations, in the two proposed architectures: without and with temporal segmentation. This study focuses experimentally on handwritten digits and signatures, but it can be extended to general handwriting synthesis.}
\label{fig:graphical_description}
\end{figure*}

Handwriting has been largely considered along the centuries as one of the most robust approaches to authenticate a person~\cite{handwritingMarcos2020}. Its application has been extended to many different authentication scenarios such as handwriting~\cite{zhang2016end}, signature~\cite{diaz2019perspective}, password~\cite{2020_TIFS_BioTouchPass2_Tolosana}, or doodles~\cite{2016_IEEE_THMS_DoodlePass_Marcos}, among others. However, a key aspect for the success of these authentication technologies is the amount of data available for subject modelling. Contrary to physiological biometric traits such as face and fingerprint, handwriting is a behavioral biometric trait, which implies that every handwritten execution is different~\cite{jain201650}. As a result, the performance of handwriting-based authentication systems can be critically affected under one-shot learning scenarios due to the high intra-subject variability~\cite{Lai_AAAI_2020}. This aspect is exacerbated by the restrictions to acquire and share handwriting biometric information for improving the learning due to legal aspects. 

This study proposes DeepWriteSYN, a novel on-line handwriting synthesis approach via deep short-term representations in order to overcome the lack of existing data in handwriting applications. Fig.~\ref{fig:graphical_description} provides a graphical representation of DeepWriteSYN. It comprises two modules: \textit{i)} an optional short-term handwriting segmentation, which divides the handwriting into short segments (e.g., individual strokes), and \textit{ii)} on-line synthesis of those short segments, which is based on deep learning technology. The main contributions of this study are:

\begin{itemize}
\item DeepWriteSYN, a novel on-line handwriting synthesiser based on a Variational Autoencoder (VAE). One of the main advantages of the approach is that the synthesis is carried out in short-time segments, being able to synthesise both given handwriting structures from unseen subjects and natural handwriting variations of given subjects. 

\item An in-depth qualitative analysis of DeepWriteSYN over two different on-line handwriting applications: signature and digit recognition, achieving good visual results.

\item A quantitative experimental framework for on-line signature verification, showing the high potential of DeepWriteSYN for improving one-shot learning scenarios.

\item To the best of our knowledge, this is the first deep synthesis approach proposed for on-line handwritten signature capable of modelling the non-linear features present in the signals as current approaches are based on the traditional Sigma LogNormal writer generation model~\cite{o2009development,ferrer2018idelog,Lai_AAAI_2020}. %(\textcolor{blue}{meter caña aqui con las limitaciones de Sigma LogNormal})

\end{itemize}

The application of DeepWriteSYN extends from the improvement of handwriting authentication scenarios using it as data augmentation technique to the generation of realistic forgeries with different quality levels~\cite{2018_HanbookBioAntiSpoofing_signature_Tolosana}, which is one of the main limitations of real authentication scenarios. In addition, it could also benefit other research lines based on time sequences such as keystroke biometrics~\cite{morales2020keystroke} or human activity/mood recognition~\cite{zhu2015naturalistic}, among many others.

%The remainder of the paper is organised as follows. Sec.~\ref{relatedWorks} summarises previous studies carried out in the synthesis of on-line handwriting via deep learning. Sec.~\ref{proposedApproach} explains the details of the proposed synthesis approach. Sec.~\ref{passwordApplications} and~\ref{signatureApplications} describe the application of the proposed synthesis approach over on-line handwritten digit and signature scenarios, respectively. Sec.~\ref{signatureVerification} describes a quantitative experimental framework showing the potential of the synthesis approach for on-line signature verification under one-shot learning scenarios. Finally, Sec.~\ref{conclusions} draws the final conclusions and points out future research lines.

\section{Related Works}\label{relatedWorks}
On-line handwriting synthesis has been extensively studied in the last decade~\cite{elarian2014handwriting}.

One of the most popular deep learning approaches in the literature is~\cite{graves2013generating}. In that study, Graves proposed the use of Recurrent Neural Networks (RNN), in particular Long Short-Term Memory (LSTM), to generate complex sequences, predicting one data point at a time. His proposed synthesis approach was tested on handwriting, achieving good visual results. 

Another important related work is~\cite{QuickDraw_2018}, where the authors presented Sketch-RNN, a sequence-to-sequence VAE trained with huge amount of data that takes in a sketch as input and outputs a synthetic sketch with different levels of variability. 

Handwriting synthesis approaches able to convert off-line images to on-line time sequences have also been proposed in the literature. In~\cite{bhunia2018handwriting,zhao2018pen}, the authors presented synthesis approaches based on the combination of Convolutional Neural Networks (CNN) and RNN/regression to predict the probability of the next stroke point position. An interesting approach was recently presented in~\cite{Cross-VAE} where the authors presented a network consisted of two VAE with a shared latent space, named Cross-VAE. Their results showed it is possible to reconstruct the images and time sequences.

On-line signature verification is undoubtedly one of the most important scenarios for the handwriting synthesis. This is motivated due to: \textit{i)} the lack of massive public databases, \textit{ii)} the low number of genuine signatures acquired per subject during the enrolment stage, and \textit{iii)} the lack of forgery signatures in real scenarios. All these aspects make it difficult to train robust biometric signature  verification systems~\cite{diaz2019perspective}. 

Currently, the Sigma LogNormal is the most popular on-line signature synthesis approach~\cite{o2009development}, although other different approaches have been proposed in the literature~\cite{galbally2012synthetic,ferrer2016behavioral}. These synthesis approaches have been successfully applied to on-line signature verification, generating synthetic samples from just one genuine signature and improving to a large extent the performance of the systems~\cite{diaz2016dynamic,Lai_AAAI_2020}. However, these synthesis approaches have an important limitation: the synthesis of the signatures are only based on linear features. This motivates DeepWriteSYN, the first deep synthesis approach proposed for on-line handwritten signature capable of modelling the non-linear features present in the handwriting signals.

%and despite the importance of synthesis techniques for on-line signature verification, none of them are based on deep learning technology, to the best of our knowledge, which is one of the main motivations of the proposed study. In addition, contrary to other deep learning approaches that require to train one model per user/class, our proposed approach is trained at stroke level and in a writer-independent mode. Therefore, a single neural network model is trained, being able to automatically generalise to other handwriting applications and unseen users.

%\begin{figure}[t]
%\begin{center}
%   \includegraphics[width=0.9\linewidth]{figure1}
%\end{center}
%   \caption{Real and synthetic examples generated using Sketch-RNN.}
%\label{fig:examples_sketches}
%\end{figure}

\section{Proposed Approach: DeepWriteSYN}\label{proposedApproach}
Fig.~\ref{fig:graphical_description} provides a graphical representation of DeepWriteSYN, the proposed on-line handwriting synthesis approach. It comprises two modules: \textit{i)} an optional temporal segmentation recommended to break long-term into short-term handwriting structures\footnote{We use short-term in our work for small handwriting segments from a fraction to full characters, doodles, or simple handwritten graphics, where higher level language or graphical structures are not present, in contrast to long-term handwriting where those language or graphical structures are present.}, and \textit{ii)} the short-term on-line handwriting synthesis implemented with a VAE. That VAE is based on the popular Sketch-RNN approach presented in~\cite{QuickDraw_2018}. Despite the good visual results achieved for sketches, it is important to highlight several limitations of that work by Ha and Eck when directly applied to handwriting: \textit{i)} Sketch-RNN does not work properly for time sequences higher than 300 samples, which is the most typical case in handwriting applications with sampling frequencies around 100-200 Hz~\cite{2017_PLOSONE_eBioSign_Tolosana}, and \textit{ii)} a specific neural network model should be trained for each specific class to achieve accurate results. 

In order to overcome these limitations, the approach presented here is based on the synthesis via short-term representations trained on datasets incorporating multiple classes, which is flexible enough to generate realistic handwriting variations of a given handwritten structure within a given population or a given subject.

\begin{figure}[t]
\begin{center}
   \includegraphics[width=0.94\linewidth]{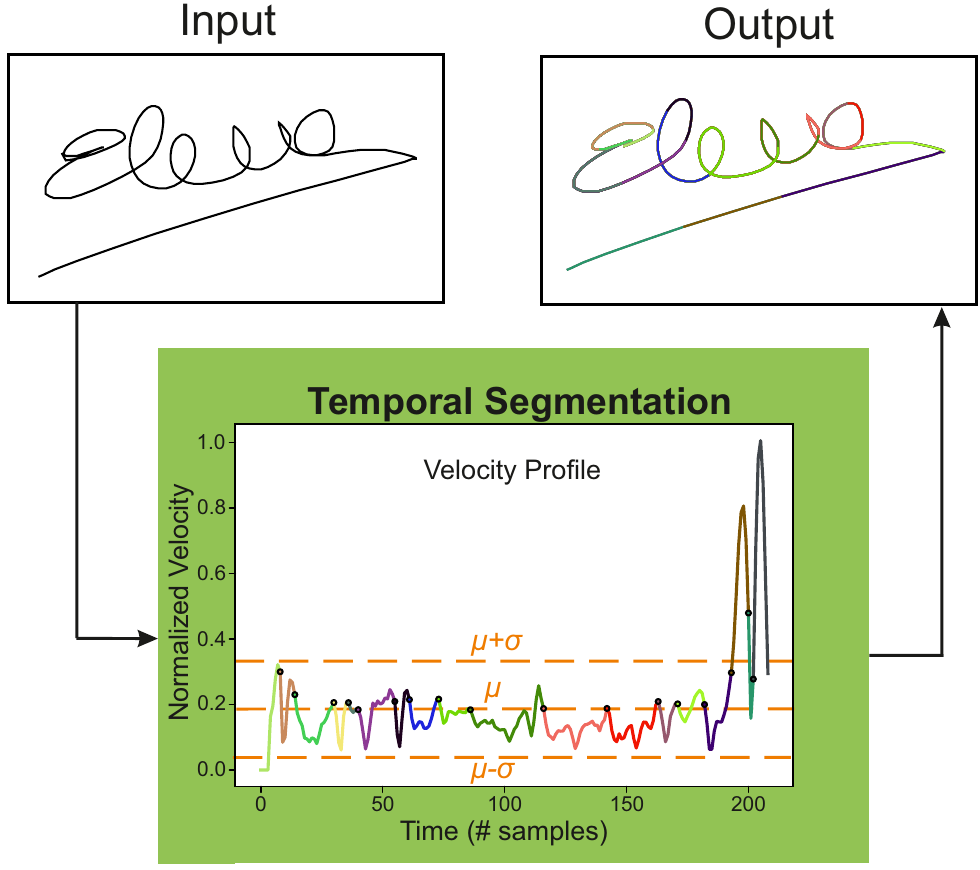}
\end{center}
   \caption{Temporal segmentation into strokes using velocity profiles. (Color image.)}
\label{fig:segmentation_module}
\end{figure}

\subsection{Temporal Segmentation}\label{StrokeSegmentation}
The temporal segmentation is an optional module considered to generate short-term representations of the handwriting. Depending on the length of the time sequences to synthesise, this module is needed to overcome the limitations of Sketch-RNN ($\leq$ 300 samples). Different segmentation approaches such as the Sigma Lognormal could be considered to extract short-term representations. For reproducibility reasons, we make use of a simple implementation based on the velocity profile $V$ of the handwriting.

%that could be further improved and optimized in future works.  

%The proposed stroke-level segmentation module is based on the velocity profile $V$ of the handwriting. 

For each handwriting information acquired, time sequences $X$ and $Y$ related to the screen coordinates of the handwriting are captured. These signals can be defined as:

\begin{equation}
\begin{array}{l}
 X =  x_1 ,x_2,\ldots,x_n,\ldots,x_N  \\
 Y = y_1 ,y_2,\ldots,y_n,\ldots,y_N  \\
 \end{array}
\end{equation}

\noindent where $N$ indicates the total number of time samples. The velocity magnitude of the writing $V=v_1,v_2,\ldots,v_n,\ldots,v_N$ can be obtained as: 

\begin{equation}
\upsilon _n  = \sqrt {\dot x_n^2  + \dot y_n^2 }
\end{equation}

%\upsilon _n  = \sqrt {(x_n-x_{n-1})^2  + (y_n-y_{n-1})^2 }

%\upsilon _n  = \sqrt {\dot x_n^2  + \dot y_n^2 }

%\upsilon(x_n,y_n)  = \sqrt {\dot x_n^2  + \dot y_n^2 }

\noindent where $\dot x_n$ and $\dot y_n$ represent the first order time derivative of consecutive samples of $x$ and $y$, respectively. Once the velocity profile of the writing is obtained, strokes are extracted similar to the approach proposed in~\cite{khan2006velocity}. The velocity profile is divided into  four regions using the mean ($\mu$) and standard deviation ($\sigma$) of the velocity profile $V$. The following three thresholds are considered: $\mu - \sigma$, $\mu$, and $\mu + \sigma$. Every time the velocity profile crosses one of the three thresholds, a stroke is extracted from the original handwriting. Fig.~\ref{fig:segmentation_module} shows an example of the proposed segmentation, including the velocity profile of the handwriting, the three thresholds considered, and the corresponding strokes extracted. As can be seen, this method allows to extract simple strokes composed of straight lines and small curves.

%what makes easier the synthesis.

%Depending on the specific handwritten application, the proposed stroke-level segmentation module may be a key aspect of the synthesis process as it makes the training process accurate and robust. This aspect is discussed in Sec.~\ref{passwordApplications} and~\ref{signatureApplications}. 

%\begin{equation}
%\begin{array}{l}
% \dot x_n =  x_n  - x_{n - 1}  \\
% \dot y_n = y_n  - y_{n - 1}  \\
% \end{array}
%\end{equation}

\subsection{Short-Term Variational Autoencoder}\label{HandwritingSynthesis}

\subsubsection{Architecture.}\label{architecture}
\begin{figure}[t]
\begin{center}
   \includegraphics[width=\linewidth]{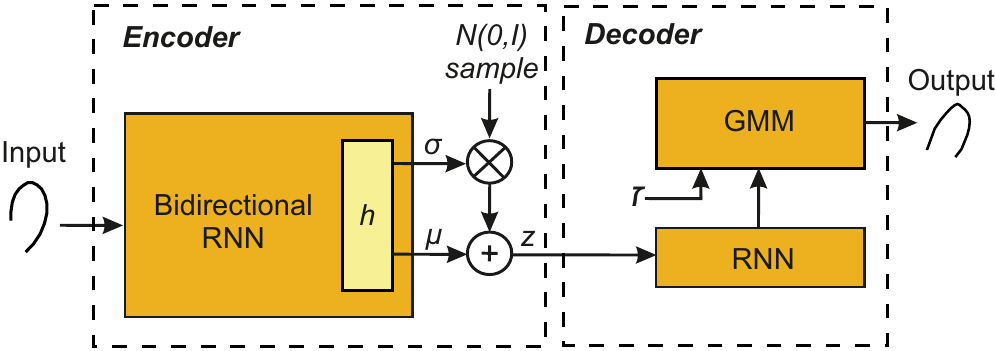}
\end{center}
   \caption{Short-Term Variational Autoencoder.}
\label{fig:synthesis_module}
\end{figure}

The short-term handwriting synthesis is based on the original Sketch-RNN presented in~\cite{QuickDraw_2018}. This is a sequence-to-sequence VAE composed of an encoder and a decoder. The source code is publicly available in GitHub\footnote{\url{https://github.com/magenta/magenta-js/tree/master/sketch}}. Fig.~\ref{fig:synthesis_module} shows the architecture of the proposed on-line handwriting synthesis module.

%The encoder is based on a Bidirectional RNN (BRNN)~\cite{2018_IEEEAccess_RNN_Tolosana}. It takes in a stroke as input, and outputs a latent feature vector $z$ of size $N_z$. Specifically, the latent feature vector $z$ is obtained following the next steps: \textit{i)} the final hidden state $h$ of the BRNN is projected into two vectors $\mu$ and $\sigma$, each of size $N_z$, using a fully-connected layer; and \textit{ii)} the vectors $\mu$ and $\sigma$, along with $N(0,I)$, a vector of IID Gaussian variables of size $N_z$, are used to construct the final latent feature vector $z$:

The encoder is based on a Bidirectional RNN (BRNN). It takes in a time sequence as input and outputs a non-deterministic latent feature vector $z$ of size $N_z$, representing the BRNN state as a Gaussian mixed with a random sample. 

Regarding the decoder, it considers an autoregressive RNN that generates output sequences conditional on a given latent feature vector $z$. The following steps are considered in the decoder: \textit{i)} first, the outputs of the decoder RNN at each time step are the parameters for a probability distribution of the next data point; and \textit{ii)} those parameters are inserted into a Gaussian Mixture Model (GMM) with $M$ normal distributions in order to finally predict the most likely data points. It is important to highlight that, similar to the encoder, the stroke output is not deterministic, but a random sequence, conditioned on the input latent feature vector $z$. The level of randomness of the output sequence can be controlled in the decoder using a temperature parameter $\tau$, which can scale the parameters of the GMM. This parameter is usually set between 0 and 1. In case $\tau = 0$, the model becomes deterministic and samples will consist of the most likely point in the probability density function.

\subsubsection{Training.}\label{training}
Both encoder and decoder are trained end-to-end following the traditional approach of the VAE~\cite{VAE_2014}, where the loss function is: 

\begin{equation}\label{eq:equation3}
Loss = L_R + w_{KL} L_{KL}
\end{equation}

\noindent being $L_R$ the Reconstruction Loss, $L_{KL}$ the Kullback-Leibler Divergence Loss, and $w_{KL}$ a weight parameter configurable during training. If $w_{KL} = 0$, the model behaves like a pure Autoencoder (AE), training to synthesise the same strokes introduced as input. Each model considered in this study is trained  from scratch.

\subsubsection{Configuration.}\label{configuration} 
We consider the same setup suggested in~\cite{QuickDraw_2018}. LSTM~\cite{2018_IEEEAccess_RNN_Tolosana} and HyperLSTM~\cite{Da_HyperNetworks_2017} are considered for the encoder and decoder RNN, respectively. Regarding the number of memory blocks, 512 are used in the encoder and 2,048 in the decoder. For the GMM, $M=20$ mixture components. The size of the latent feature vector $N_z$ is 128. During training, layer normalization and recurrent dropout with a probability of 90\% are considered. Adam optimiser is considered with default parameters (learning rate of 0.0001).

\begin{figure*}[t]
\begin{center}
   \includegraphics[width=0.8\linewidth]{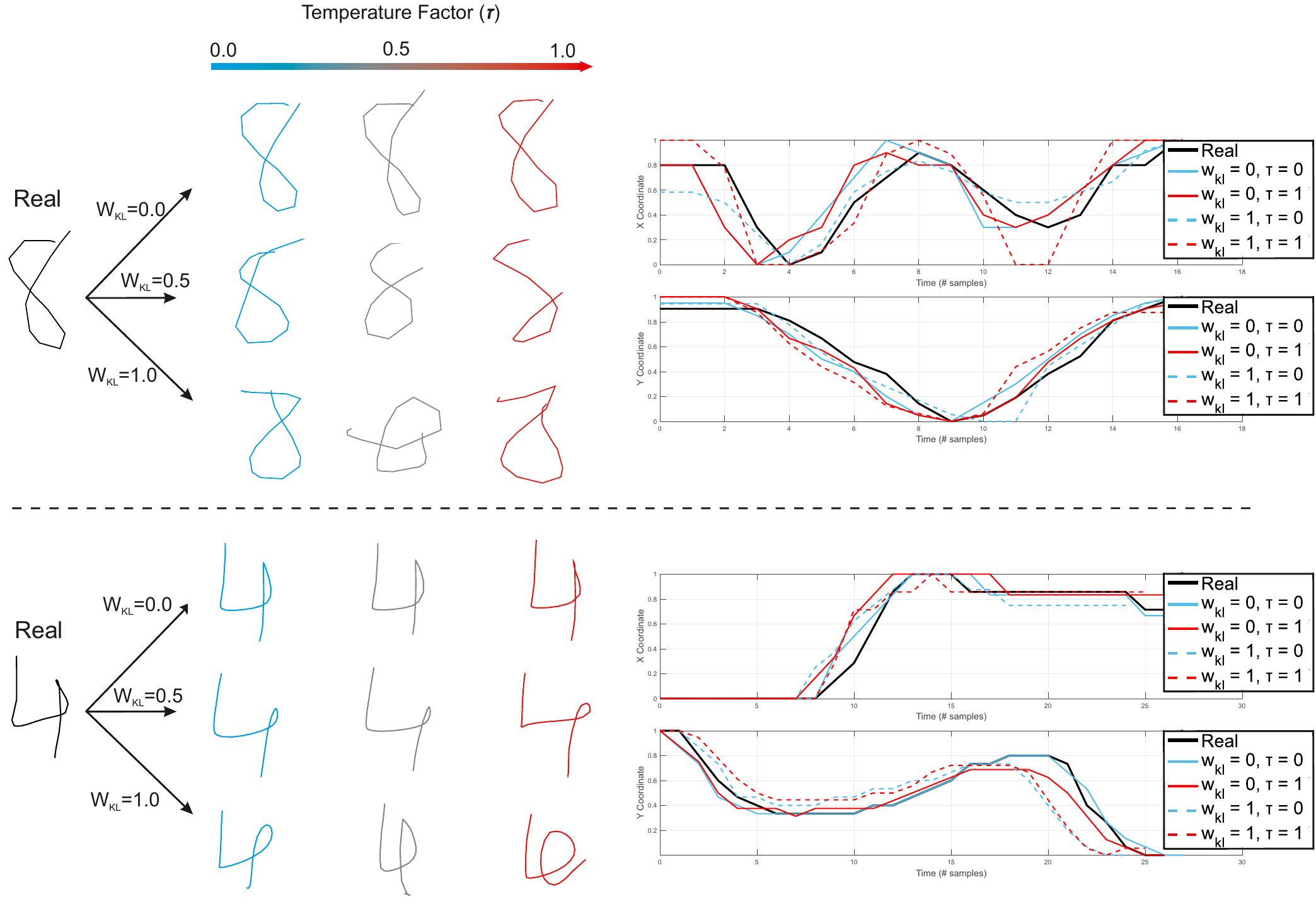}
\end{center}
   \caption{Images and temporal sequences (\textit{X} and \textit{Y} coordinates) of the real (solid black) and synthetic digits (blue and red).}
\label{fig:examples_digits}
\end{figure*}

\begin{figure}[t]
\centering
\subfigure[Real]{\label{tsne_real}
\includegraphics[width=0.47\linewidth]{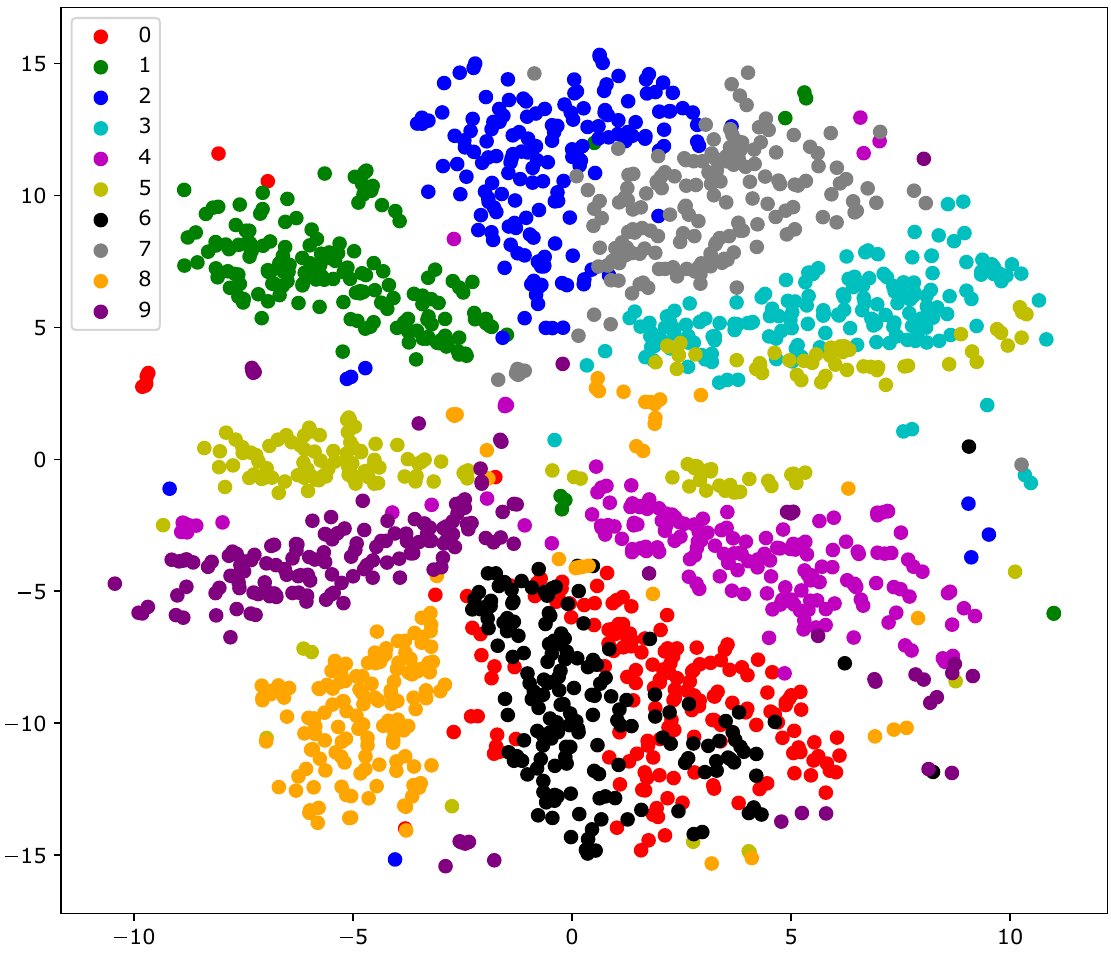}}
\subfigure[Synthetic]{\label{tsne_synthetic}
\includegraphics[width=0.47\linewidth]{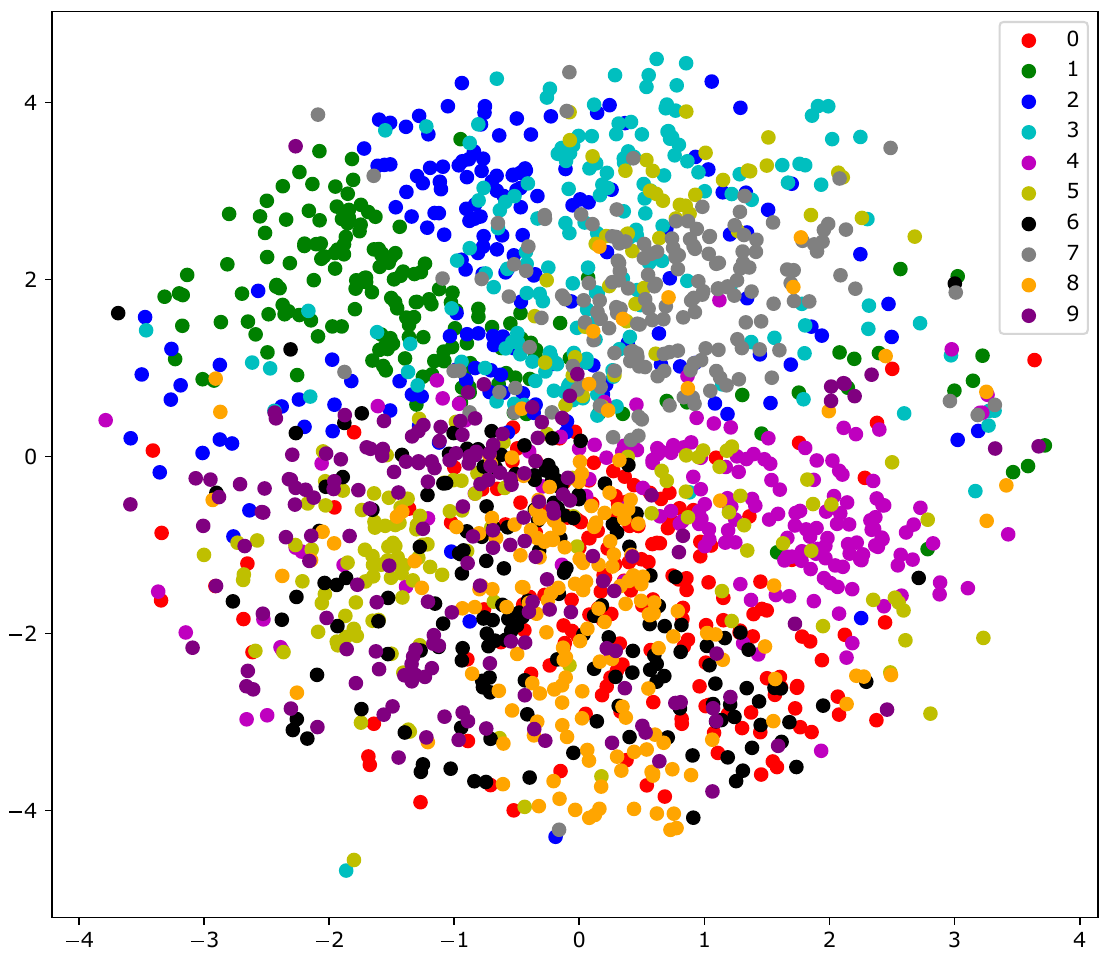}}
\caption{t-SNE distributions of real and synthetic handwritten digits generated using DeepWriteSYN. (Color image.)} 
\label{fig:tsne_digits}
\end{figure}

\section{Digit Synthesis}\label{passwordApplications}
We first analyse in this section DeepWriteSYN over an initial qualitative example: synthesis of handwritten digits from 0 to 9. In this scenario: \textit{i)} the number of classes to synthesise is fixed to ten (0-9 digits), and \textit{ii)} the number of time samples of each handwritten digit is always $\leq$ 100 and there are no long-term handwriting structures in place. Therefore, we process this case purely short-term without temporal segmentation, as indicated in the upper half of Fig.~\ref{fig:graphical_description}.

\subsection{Experimental Protocol}\label{passwordExperimentalProtocol}
DeepWriteSYN is trained from scratch using handwritten digits of the public eBioDigitDB database\footnote{\url{https://github.com/BiDAlab/eBioDigitDB}} \cite{2019_TMC_BioTouchPass_Tolosana}. This database comprises on-line handwritten digits from 0 to 9 acquired using 93 total subjects. Handwritten digits were captured using the finger over a Samsung Galaxy Note 10.1 device. Two different acquisition sessions are considered with a time gap of at least three weeks between them.

In this experimental framework, eBioDigitDB is divided into development and evaluation datasets, which comprise different subjects. The development dataset is considered in the training process, using 6,200 total samples (620 samples per digit). Finally, after training, we consider the unseen subjects included in the evaluation. This evaluation dataset comprises 1,200 total samples (120 samples per digit).

\subsection{Results}\label{passwordResults}
The following two parameters of the short-term VAE are configured to control the variability of the generated digits:

\begin{itemize}
\item $\tau$: the temperature parameter that can scale the values of the GMM, controlling the randomness of the output sequence.
\item $w_{KL}$: the weight parameter of the loss function that controls the relation between $L_R$ and $L_{KL}$, see Eq.~\ref{eq:equation3}.
\end{itemize}

\begin{figure*}[t]
\begin{center}
   \includegraphics[width=0.81\linewidth]{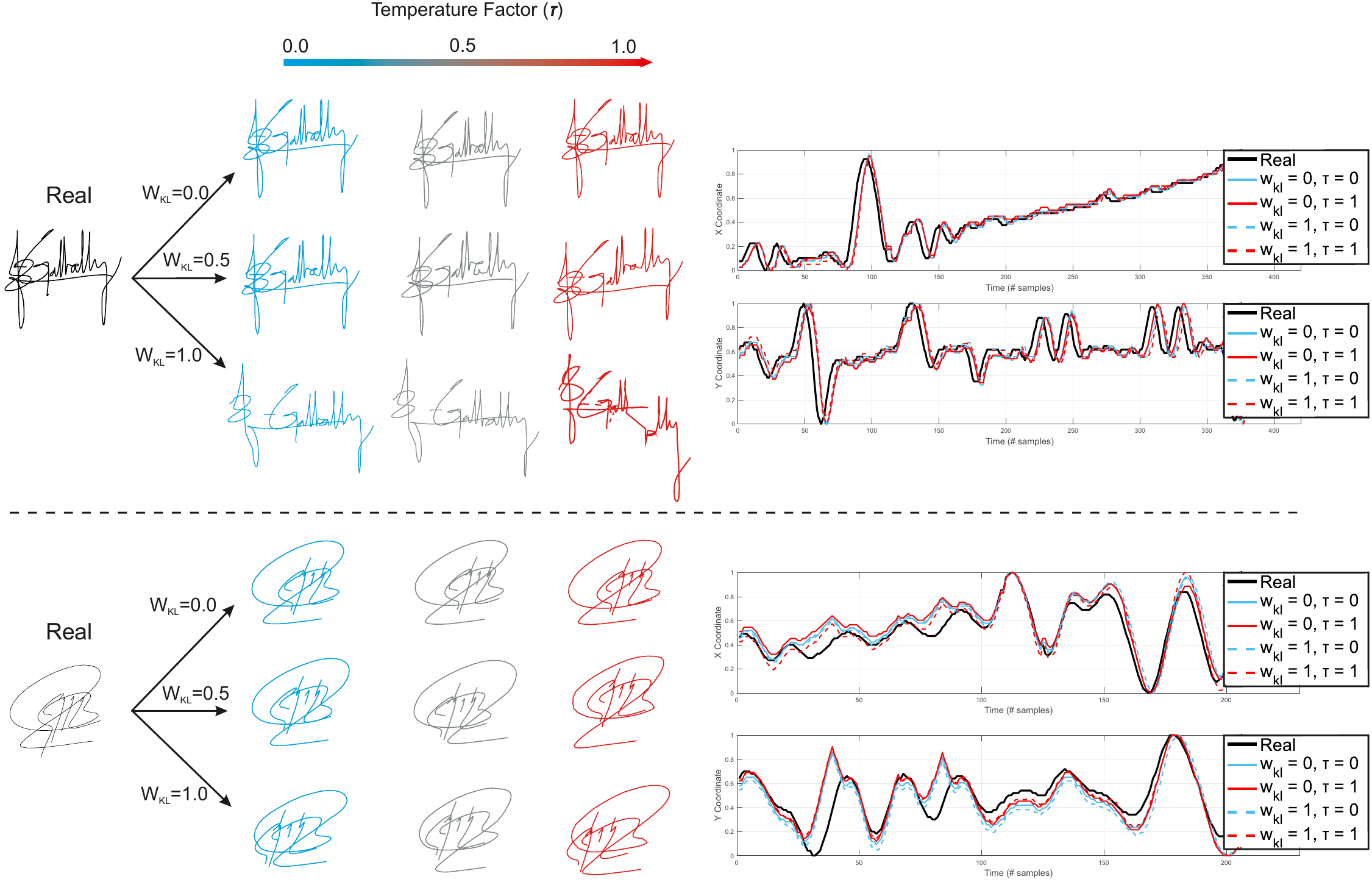}
\end{center}
   \caption{Images and temporal sequences (\textit{X} and \textit{Y} coordinates) of the real (solid black) and synthetic signatures (blue and red). }
\label{fig:examples_signatures}
\end{figure*}

Fig.~\ref{fig:examples_digits} shows the effect of $w_{KL}$ and $\tau$ parameters in the synthesis process. Both images and time sequences (\textit{X} and \textit{Y} coordinates) are included for completeness.

Analysing the effect of the $w_{KL}$ parameter for a fixed $\tau=0$, very similar results are obtained for the Autoencoder approach (i.e., $w_{KL}=0$) compared with the real samples. When the value of $w_{KL}$ increases, a higher variability of the synthetic samples is observed. For example, generating samples with different geometrical aspects of the loops, and also with strokes longer/shorter in time.

The variability observed in the synthetic strokes is even higher when we increase the values of the $w_{KL}$ and $\tau$ parameters. However, in some cases, values of $w_{KL}$ and $\tau$ close to 1 can distort the input digit as shown in Fig.~\ref{fig:examples_digits}. A combination of different $w_{KL}$ and $\tau$ parameters for each stroke could further modify the intra-subject variability of the samples.

Finally, to provide further insights, we show in Fig.~\ref{fig:tsne_digits} the t-Distributed Stochastic Neighbor Embedding (t-SNE) feature distributions of real and synthetic handwritten digits generated using DeepWriteSYN. As can be seen, the proposed synthesis approach (in Fig. \ref{tsne_synthetic}, $w_{KL}=0.25$, $\tau=0$) is able to fill more densely the feature space following approximately the same class distributions and relations of the original feature space.

\section{Signature Synthesis}\label{signatureApplications}
This section analyses qualitatively the potential of DeepWriteSYN over a real scenario: synthesis of handwritten signatures. This scenario is more challenging than the previous one (digit synthesis) due to: \textit{i)} the number of classes to synthesise is unlimited as each of us perform a unique signature, and \textit{ii)} signatures may contain complex handwriting structures such as concatenated characters, words, graphical embellishment, etc. that the learning structure (Sketch-RNN) may not represent well. Therefore, we process this case as long-term including temporal segmentation as indicated in the lower half of Fig.~\ref{fig:graphical_description}, so our core VAE works with short-time segments consisting of simple handwriting structures. 

\subsection{Experimental Protocol}\label{signatureExperimentalProtocol}
DeepWriteSYN is trained from scratch in this case using signatures from the public DeepSignDB\footnote{\url{https://github.com/BiDAlab/DeepSignDB}} \cite{2020_Arxiv_DeepSign_Tolosana}. This database comprises 1,526 total subjects divided into development (1,084 subjects) and evaluation (442 subjects). For each subject, genuine signatures were acquired in multiple sessions with different time gaps between them. Also, signatures were acquired using different devices and writing inputs. In this experimental framework, the development dataset of DeepSignDB is only considered to train DeepWriteSYN. Finally, after training, we consider the unseen subjects included in the evaluation dataset to see the ability of the proposed approach to synthesise realistic new realizations of a given signature  from a subject unseen in the learning stage. Only those signatures acquired through the stylus are considered in this experiment.

%As a result, we use 12,782 genuine signatures

%This evaluation dataset comprises 11,752 signatures. 

\subsection{Results}\label{signatureResults}
Fig.~\ref{fig:examples_signatures} illustrates the effect of DeepWriteSYN for on-line handwritten signature regarding the two configuration parameters $\tau$ and $w_{KL}$. Both images and time sequences (\textit{X} and \textit{Y} coordinates) are included for completeness.

\begin{figure}[t]
\centering
\subfigure{\label{fig:exp1_parameters}
\includegraphics[width=0.745\linewidth]{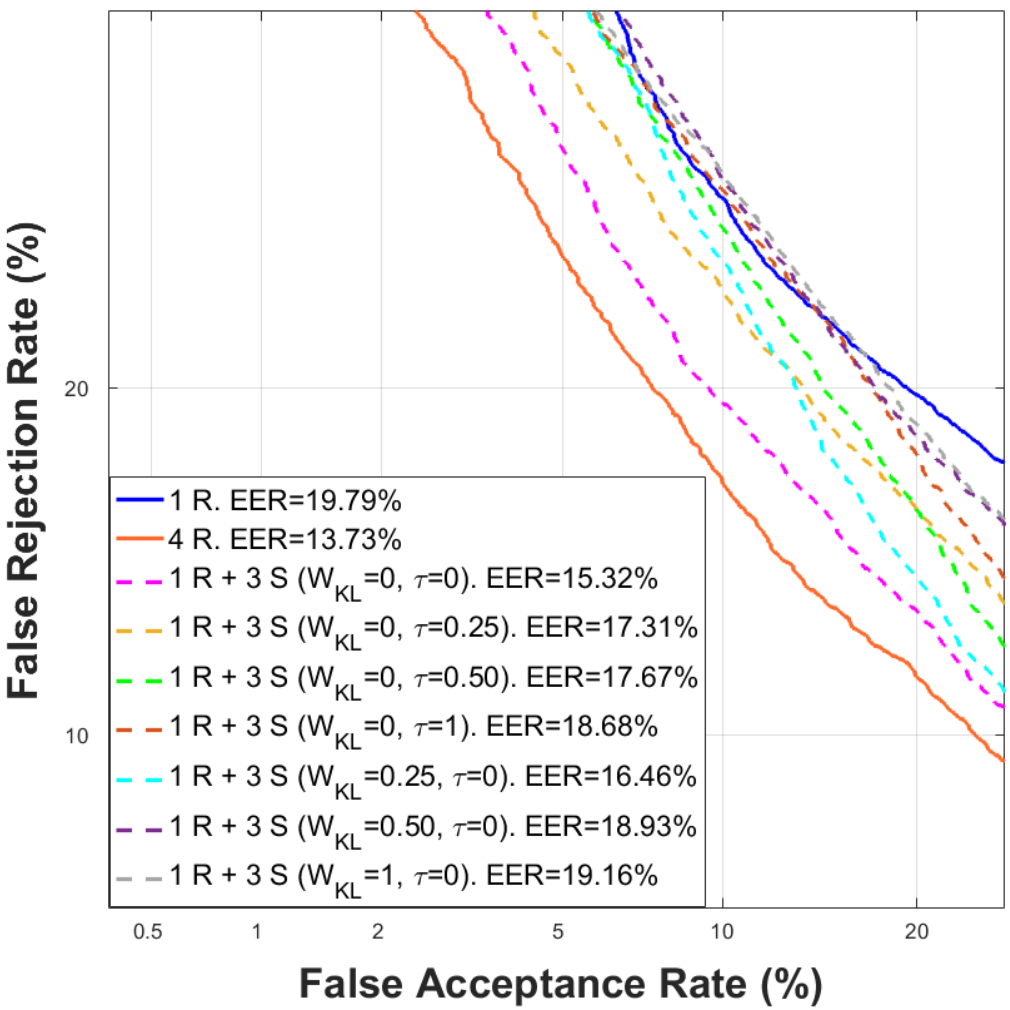}}
\subfigure{\label{fig:exp2_parameters}
\includegraphics[width=0.745\linewidth]{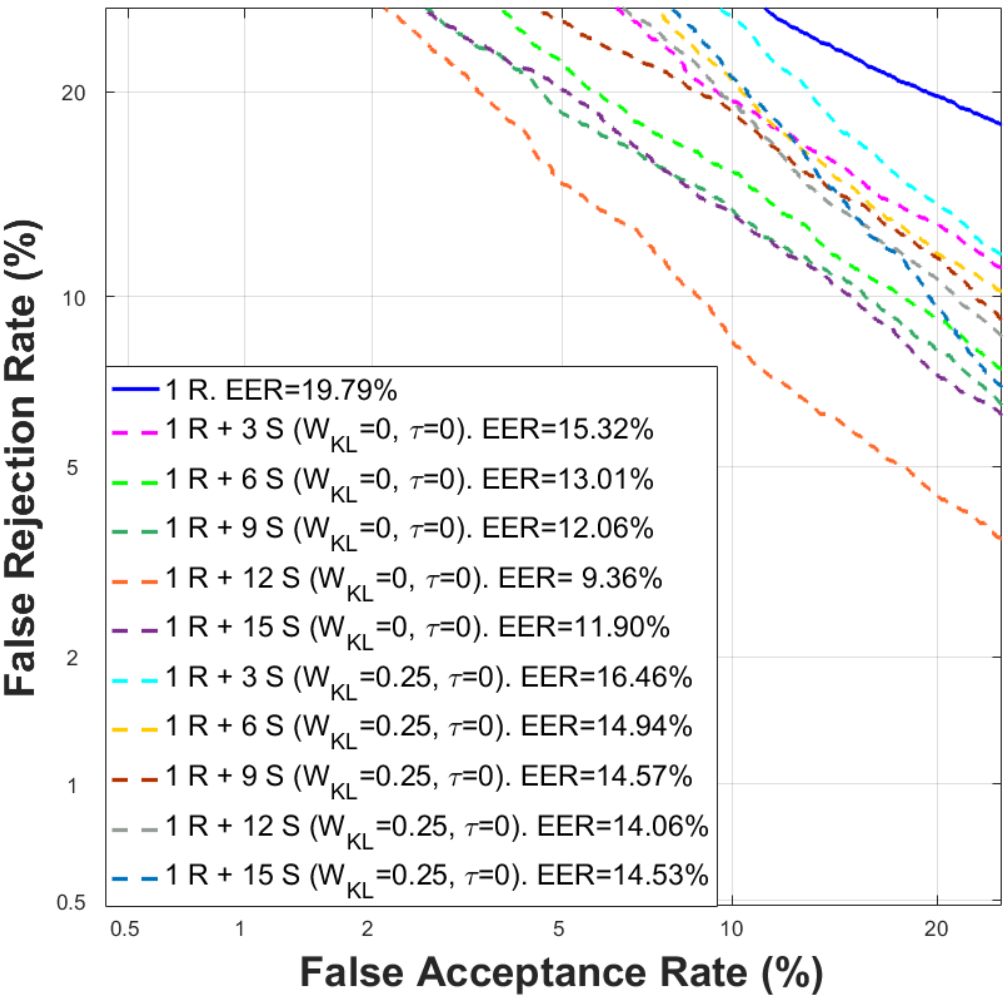}}
\caption{DET curves and EER (\%) results on the evaluation dataset regarding \textbf{(Top, Exp. 1)} the different $w_{KL}$ and $\tau$ values of the proposed synthesis approach, and \textbf{(Bottom, Exp. 2)} the different number of synthetic samples generated. } 
\label{fig:det_curves}
\end{figure}

We first analyse the effect of the $w_{KL}$ parameter in the synthesis process, considering $\tau=0$. When $w_{KL} = 0$, the proposed approach behaves like an Autoencoder, generating very similar samples compared with the real ones, proving the ability of the model to generalise to strokes unseen during the learning stage. If we increase the value of $w_{KL}$ closer to 1, a higher variability is included in the synthetic samples. For example, generating strokes longer/shorter in time (see the right/left movements of the time sequences along the horizontal axis), and also with different geometrical shapes, especially in the embellishment of the signature (see the bottom signature of Fig.~\ref{fig:examples_signatures}) and closed letters (see the \textit{a}, \textit{l}, and \textit{y} of the top signature). 

Analysing the effect of the temperature parameter $\tau$ for a specific value of $w_{KL}$, the higher the value of $\tau$ is, the higher is the resulting variability in the synthesis. For example, shifting the spatial position of the strokes (see the variability of the \textit{X} and \textit{Y} coordinates in the vertical axis of the time sequences) and also enlarging/shortening the strokes.

Finally, a combination of $w_{KL}$ and $\tau$ parameters can further modify the synthetic signature compared with the real one, simulating the typical and natural intra-subject variability. However, in some cases, values of $w_{KL}$ and $\tau$ close to 1 can distort the input signature as shown in the top signature of Fig.~\ref{fig:examples_signatures}. It is important to highlight that in these examples the same $w_{KL}$ and $\tau$ parameters are applied over the whole signature. Nevertheless, different parameters could be considered for each temporal segment of the signature, controlling in a time-dependent way the intra-subject variability.

\section{Application: Improving One-Shot Learning in Signature Verification}\label{signatureVerification}
This section analyses quantitatively the potential of DeepWriteSYN for handwriting scenarios. In particular, we focus on the popular task of on-line signature verification~\cite{diaz2019perspective}, considering 
one-shot learning scenarios, i.e., just one enrolment signature is available per subject.

\subsection{Experimental Protocol}\label{ExperimentalProtocolVerification}
We consider as baseline the state-of-the-art on-line signature verification system presented in~\cite{2018_IEEEAccess_RNN_Tolosana}. This system is based on BRNN with a Siamese architecture.

The BiosecurID dataset included in the public DeepSignDB database\footnote{\url{https://github.com/BiDAlab/DeepSignDB}}~\cite{2020_Arxiv_DeepSign_Tolosana} is considered in this experimental framework. We consider the same experimental protocol proposed by the authors of DeepSignDB: 268 subjects for development and 132 remaining subjects for the final evaluation. Also, we focus on the most challenging impostor scenario, skilled forgeries, as the forgers had access to the image and dynamics of the signatures to forge and practise as many times as they wanted.

It is important to remark that DeepWriteSYN is only considered in the development stage as a data augmentation technique for improving the enrollment of the tested subjects. For the final evaluation of the signature verification system, the same signature comparison files provided in~\cite{2020_Arxiv_DeepSign_Tolosana} are considered here in order to perform a fair and reproducible experimental framework.

%It is important to highlight that different users are considered in each dataset to avoid biased results. In order to simulate one-shot learning scenarios, only one enrolment signature is considered per user in order to train the on-line signature verification system

%\begin{figure}[t]
%\begin{center}
%   \includegraphics[width=0.6\linewidth]{DET_exp1}
%\end{center}
%   \caption{\textbf{Exp. 1:} DET curves and EER (\%) results on the evaluation dataset regarding the different $w_{KL}$ and $\tau$ values of the proposed synthesis approach.}
%\label{fig:exp1_parameters}
%\end{figure} 
%
%
%\begin{figure}[t]
%\begin{center}
%   \includegraphics[width=0.6\linewidth]{DET_exp2}
%\end{center}
%   \caption{\textbf{Exp. 2:} DET curves and EER (\%) results on the evaluation dataset regarding the different number of synthetic samples generated.}
%\label{fig:exp2_parameters}
%\end{figure} 

\subsection{Exp 1. Analysis of the Synthesis Parameters}\label{Experiment1}
This section analyses the effect of the synthesis parameters $\tau$ and $w_{KL}$ on the system performance of the on-line signature verification system. Fig.~\ref{fig:det_curves} (Top) shows the Detection Error Tradeoff (DET) curve together with the Equal Error Rates (EERs) achieved on the final evaluation dataset for the different experiments conducted. For the baseline (solid blue curve), we consider the case of using just one real enrolment signature per subject, simulating traditional scenarios. Then, we test DeepWriteSYN (dashed curves), generating from the original enrolment sample 3 more synthetic samples using different $\tau$ and $w_{KL}$ parameters. For completeness, we also include the ideal scenario of having 4 real enrolment signatures per subject (solid orange curve).

First, we can see in Fig.~\ref{fig:det_curves} (Top) that all experiments based on DeepWriteSYN outperform the baseline result (19.79\% EER). In particular, the best results are achieved for the $w_{KL}$ values 0 and 0.25 when $\tau=0$, with EER results of 15.32\% and 16.46\%, respectively. This is a relative improvement of up to 23\% EER compared with the baseline results. In addition, the results achieved are very similar compared with the ideal case of having 4 real signatures (13.73\% EER), proving the success of DeepWriteSYN.

\subsection{Exp 2. Analysis of the \# Synthetic Samples}\label{Experiment2}
This section analyses how the number of synthetic samples generated affects the system performance of the on-line signature verification system. Fig.~\ref{fig:det_curves} (Bottom) shows the DET curve and EER (\%) results achieved on the final evaluation dataset. Regarding the synthesis parameters, we consider the $w_{KL}$ and $\tau$ values that achieved the best results in the previous experiment (i.e., $w_{KL}=0/0.25$ and $\tau=0$).

In general, a system performance improvement is observed when increasing the number of synthetic samples (up to 12). This improvement is especially considerable for $w_{KL}=0$, with a 9.36\% EER when using 12 synthetic samples. This is a relative improvement of more than 50\% EER compared with the baseline result (19.79\% EER), proving the success of DeepWriteSYN to improve already trained deep learning systems over one-shot learning scenarios.

Finally, we compare the skilled forgery results achieved with the state of the art~\cite{Lai_AAAI_2020}. In that work, the traditional Sigma LogNormal model was considered to generate synthetic samples to improve one-shot learning scenarios. Signature verification systems based on deep learning were considered, achieving relative improvements of around 7-14\% EER for the scenario of using one real signature and 15 synthetic. Our DeepWriteSYN approach presented here achieves relative improvements higher than 50\% EER with fewer synthetic signatures needed ($\leq$12).

%Finally, we compare the skilled forgery results achieved with the state of the art~\cite{diaz2016dynamic,Lai_AAAI_2020}. In both of them, the traditional Sigma LogNormal model was considered to generate synthetic samples to better train one-shot learning scenarios. In~\cite{diaz2016dynamic}, the authors achieved relative improvements higher than 50\% EER for a traditional DTW system. However, around 64-128 synthetic signatures were needed, being a very time consuming process. In~\cite{Lai_AAAI_2020}, the authors studied the effects over recent deep learning models. Their proposed approach achieved relative improvements of around 7-14\% EER for the scenario of using one real signature and 15 synthetic. The work presented here based on DeepWriteSYN achieves relative improvements higher than 50\% EER with fewer synthetic signatures needed ($\leq$12).

%14.07\% and 7.74\% EERs for the MCYT-100

\section{Limitations and Future Work}\label{limitations}
Some aspects of DeepWriteSYN are improvable, such as the segmentation of the handwriting into short-term representations, and the final reconstruction of the output sequence. Currently, each synthetic short-term representation is concatenated to form the long-term signature, following the same order than the original signature, which is not always realistic. To improve that, our short-term method may be combined with other strategies able to model better the long-term dependencies in handwriting via deep learning (like the classical work by Graves) or stochastic methods.

For future work we will also investigate: \textit{i)} the configuration parameters regarding the handwriting complexity~\cite{2019_Arxiv_ComplexityPenTouch_Tolosana}, \textit{ii)} the synthesis of realistic forgeries with different qualities~\cite{2018_HanbookBioAntiSpoofing_signature_Tolosana}, and \textit{iii)} the application to other research lines based on time sequences such as keystroke biometrics~\cite{morales2020keystroke}, and human activity recognition~\cite{zhu2015naturalistic}. 

\section{Conclusions}\label{conclusions}
This study has presented DeepWriteSYN, a novel on-line handwriting synthesiser via deep short-term representations. One of the main advantages of the approach is that the synthesis is carried out in short segments, being able to synthesise general handwriting even from unseen subjects. 

We have performed an in-depth analysis of DeepWriteSYN over two different on-line handwriting scenarios: signatures and digits, achieving good visual results. Also, we have carried out a quantitative experimental framework for on-line signature verification showing the high potential of DeepWriteSYN for challenging one-shot learning scenarios.

%, with a relative improvement higher than 50\% EER, proving the success of the proposed synthesis approach.

\section*{Acknowledgments}
This work has been supported by projects: PRIMA (H2020-MSCA-ITN-2019-860315), TRESPASS-ETN (H2020-MSCA-ITN-2019-860813), BIBECA (MINECO/FEDER RTI2018-101248-B-I00), Bio-Guard (Ayudas Fundaci\'on BBVA 2017) and by UAM-Cecabank. Ruben Tolosana is supported by Consejer\'ia de Educaci\'on, Juventud y Deporte de la Comunidad de Madrid y Fondo Social Europeo.

\bibliography{egbib2}

\end{document}